\documentclass[12pt]{elsarticle}

\usepackage{lineno}
\usepackage{url}

\usepackage{booktabs}
\usepackage{framed}
\usepackage{scrextend}
\usepackage{bm}
\usepackage[mathscr]{eucal}
\usepackage{amsmath}
\usepackage{txfonts}
\usepackage{multirow}
\usepackage{changepage}
\usepackage{url}
\usepackage{graphicx}

\usepackage{color,soul}

\journal{Information Processing \& Management}

\begin{document}

\begin{frontmatter}

\title{Discourse-Aware Rumour Stance Classification in Social Media Using Sequential Classifiers}

\author[warwick]{Arkaitz Zubiaga}
\cortext[mycorrespondingauthor]{Corresponding author}
\ead{a.zubiaga@warwick.ac.uk}

\author[warwick,ati]{Elena Kochkina}
\author[warwick,ati]{Maria Liakata}
\author[warwick,ati]{Rob Procter}
\author[sheffield]{Michal Lukasik}
\author[sheffield]{Kalina Bontcheva}
\author[melbourne]{Trevor Cohn}
\author[ucph]{Isabelle Augenstein}

\address[warwick]{University of Warwick, Coventry, UK}
\address[ati]{Alan Turing Institute, London, UK}
\address[sheffield]{University of Sheffield, Sheffield, UK}
\address[melbourne]{University of Melbourne, Melbourne, Australia}
\address[ucph]{University of Copenhagen, Copenhagen, Denmark}

\begin{abstract}
 Rumour stance classification, defined as classifying the stance of specific social media posts into one of supporting, denying, querying or commenting on an earlier post, is becoming of increasing interest to researchers. While most previous work has focused on using individual tweets as classifier inputs, here we report on the performance of sequential classifiers that exploit the discourse features inherent in social media interactions or `conversational threads'. Testing the effectiveness of four sequential classifiers -- Hawkes Processes, Linear-Chain Conditional Random Fields (Linear CRF), Tree-Structured Conditional Random Fields (Tree CRF) and Long Short Term Memory networks (LSTM) -- on eight datasets associated with breaking news stories, and looking at different types of local and contextual features, our work sheds new light on the development of accurate stance classifiers. We show that sequential classifiers that exploit the use of discourse properties in social media conversations while using only local features, outperform non-sequential classifiers. Furthermore, we show that LSTM using a reduced set of features can outperform the other sequential classifiers; this performance is consistent across datasets and across types of stances. To conclude, our work also analyses the different features under study, identifying those that best help characterise and distinguish between stances, such as supporting tweets being more likely to be accompanied by evidence than denying tweets. We also set forth a number of directions for future research.
\end{abstract}


\end{frontmatter}

\section{Introduction}

Social media platforms have established themselves as important sources for learning about the latest developments in breaking news. People increasingly use social media for news consumption \cite{hermida2012share,mitchell2015millennials,zubiaga2015real}, while media professionals, such as journalists, increasingly turn to social media for news gathering \cite{zubiaga2013curating} and for gathering potentially exclusive updates from eyewitnesses \cite{diakopoulos2012finding,tolmie2017microblog}. Social media platforms such as Twitter are a fertile and prolific source of breaking news, occasionally even outpacing traditional news media organisations \cite{kwak2010twitter}. This has led to the development of multiple data mining applications for mining and discovering events and news from social media \cite{dong2015multiscale,stilo2016efficient}. However, the use of social media also comes with the caveat that some of the reports are necessarily rumours at the time of posting, as they have yet to be corroborated and verified \cite{mendoza2010twitter,procter2013readingb,procter2013readinga}. The presence of rumours in social media has hence provoked a growing interest among researchers for devising ways to determine veracity in order to avoid the diffusion of misinformation \cite{derczynski2015pheme}.

Resolving the veracity of social rumours requires the development of a rumour classification system and we described in \cite{zubiaga2017detection}, a candidate architecture for such a system consisting of the following four components: (1) detection, where emerging rumours are identified, (2) tracking, where those rumours are monitored to collect new related tweets, (3) stance classification, where the views expressed by different tweet authors are classified, and (4) veracity classification, where knowledge garnered from the stance classifier is put together to determine the likely veracity of a rumour.

In this work we focus on the development of the third component, a stance classification system, which is crucial to subsequently determining the veracity of the underlying rumour. The stance classification task consists in determining how individual posts in social media observably orientate to the postings of others \cite{walker2012stance,qazvinian2011rumor}. For instance, a post replying with ``no, that's definitely false'' is \textit{denying} the preceding claim, whereas ``yes, you're right'' is \textit{supporting} it. It has been argued that aggregation of the distinct stances evident in the multiple tweets discussing a rumour could help in determining its likely veracity, providing, for example, the means to flag highly disputed rumours as being potentially false \cite{mendoza2010twitter}. This approach has been justified by recent research that has suggested that the aggregation of the different stances expressed by users can be used for determining the veracity of a rumour \cite{derczynski2015pheme,liu2015realtime}.

In this work we examine in depth the use of so-called sequential approaches to the rumour stance classification task. Sequential classifiers are able to utilise the discursive nature of social media \cite{tolmie2017microblog}, learning from how `conversational threads' evolve for a more accurate classification of the stance of each tweet. The use of sequential classifiers to model the conversational properties inherent in social media threads is still in its infancy. For example, in preliminary work we showed that a sequential classifier modelling the temporal sequence of tweets outperforms standard classifiers \cite{lukasik2016hawkes,zubiaga2016coling}. Here we extend this preliminary experimentation in four different directions that enable exploring further the stance classification task using sequential classifiers: (1) we perform a comparison of a range of sequential classifiers, including a Hawkes Process classifier, a Linear CRF, a Tree CRF and an LSTM; (2) departing from the use of only local features in our previous work, we also test the utility of contextual features to model the conversational structure of Twitter threads; (3) we perform a more exhaustive analysis of the results looking into the impact of different datasets and the depth of the replies in the conversations on the classifiers' performance, as well as performing an error analysis; and (4) we perform an analysis of features that gives insight into what characterises the different kinds of stances observed around rumours in social media. To the best of our knowledge, dialogical structures in Twitter have not been studied in detail before for classifying each of the underlying tweets and our work is the first to evaluate it exhaustively for stance classification. Twitter conversational threads are identifiable by the relational features that emerge as users respond to each others' postings, leading to tree-structured interactions. The motivation behind the use of these dialogical structures for determining stance is that users' opinions are expressed and evolve in a discursive manner, and that they are shaped by the interactions with other users.

The work presented here advances research in rumour stance classification by performing an exhaustive analysis of different approaches to this task. In particular, we make the following contributions:

\begin{itemize}
\item We perform an analysis of whether and the extent to which use of the sequential structure of conversational threads can improve stance classification in comparison to a classifier that determines a tweet's stance from the tweet in isolation. To do so, we evaluate the effectiveness of a range of sequential classifiers: (1) a state-of-the-art classifier that uses Hawkes Processes to model the temporal sequence of tweets \cite{lukasik2016hawkes}; (2) two different variants of Conditional Random Fields (CRF), i.e., a linear-chain CRF and a tree CRF; and (3) a classifier based on Long Short Term Memory (LSTM) networks. We compare the performance of these sequential classifiers with non-sequential baselines, including the non-sequential equivalent of CRF, a Maximum Entropy classifier.

\item We perform a detailed analysis of the results broken down by dataset and by depth of tweet in the thread, as well as an error analysis to further understand the performance of the different classifiers. We complete our analysis of results by delving into the features, and exploring whether and the extent to which they help characterise the different types of stances.
\end{itemize}

Our results show that sequential approaches do perform substantially better in terms of macro-averaged F1 score, proving that exploiting the dialogical structure improves classification performance. Specifically, the LSTM achieves the best performance in terms of macro-averaged F1 scores, with a performance that is largely consistent across different datasets and different types of stances. Our experiments show that LSTM performs especially well when only local features are used, as compared to the rest of the classifiers, which need to exploit contextual features to achieve comparable -- yet still inferior -- performance scores. Our findings reinforce the importance of leveraging conversational context in stance classification. Our research also sheds light on open research questions that we suggest should be addressed in future work. Our work here complements other components of a rumour classification system that we implemented in the PHEME project, including a rumour detection component \cite{zubiaga2016learning,zubiaga2017exploiting}, as well as a study into the diffusion of and reactions to rumour \cite{zubiaga2016analysing}.

\section{Related Work}

Stance classification is applied in a number of different scenarios and domains, usually aiming to classify stances as one of ``in favour'' or ``against''. This task has been studied in political debates \cite{hasan2013extra,walker2012your}, in arguments in online fora \cite{hasan2013stance,sridhar2014collective} and in attitudes towards topics of political significance \cite{augenstein2016stance,mohammad2016semeval,Augenstein2016SemEval}. In work that is closer to our objectives, stance classification has also been used to help determine the veracity of information in micro-posts \cite{qazvinian2011rumor}, often referred to as \textit{rumour stance classification} \cite{lukasik2015classifying,lukasik2016hawkes,procter2013readinga,zubiaga2016coling}. The idea behind this task is that the aggregation of distinct stances expressed by users in social media can be used to assist in deciding if a report is actually true or false \cite{derczynski2015pheme}. This may be particularly useful in the context of rumours emerging during breaking news stories, where reports are released piecemeal and which may be lacking authoritative review; in consequence, using the `wisdom of the crowd' may provide a viable, alternative approach. The types of stances observed while rumours circulate, however, tend to differ from the original ``in favour/against'', and different types of stances have been discussed in the literature, as we review next.

Rumour stance classification of tweets was introduced in early work by Qazvinian et al. \cite{qazvinian2011rumor}. The line of research initiated by \cite{qazvinian2011rumor} has progressed substantially with revised definitions of the task and hence the task tackled in this paper differs from this early work in a number of aspects. Qazvinian et al. \cite{qazvinian2011rumor} performed 2-way classification of each tweet as \textit{supporting} or \textit{denying} a long-standing rumour such as disputed beliefs that \textit{Barack Obama is reportedly Muslim}. The authors used tweets observed in the past to train a classifier, which was then applied to new tweets discussing the same rumour. In recent work, rule-based methods have been proposed as a way of improving on Qazvinian et al.'s baseline method; however, rule-based methods are likely to be difficult -- if not impossible -- to generalise to new, unseen rumours. Hamidian et al. \cite{hamidian2016rumor} extended that work to analyse the extent to which a model trained from historical tweets could be used for classifying new tweets discussing the same rumour.

The work we present here has three different objectives towards improving stance classification. First, we aim to classify the stance of tweets towards rumours that emerge while breaking news stories unfold; these rumours are unlikely to have been observed before and hence rumours from previously observed events, which are likely to diverge, need to be used for training. As far as we know, only work by Lukasik et al.  \cite{lukasik2015classifying,lukasik2016using,lukasik2016hawkes} has tackled stance classification in the context of breaking news stories applied to new rumours. Zeng et al. \cite{zeng2016unconfirmed} have also performed stance classification for rumours around breaking news stories, but overlapping rumours were used for training and testing. Augenstein et al. \cite{augenstein2016stance,Augenstein2016SemEval} studied stance classification of unseen events in tweets, but ignored the conversational structure. Second, recent research has proposed that a 4-way classification is needed to encompass responses seen in breaking news stories \cite{procter2013readinga,zubiaga2016analysing}. Moving away from the 2-way classification above, which \cite{procter2013readinga} found to be limited in the context of rumours during breaking news, we adopt this expanded scheme to include tweets that are \textit{supporting}, \textit{denying}, \textit{querying} or \textit{commenting} rumours. This adds more categories to the scheme used in early work, where tweets would only support or deny a rumour, or where a distinction between querying and commenting is not made \cite{augenstein2016stance,mohammad2016semeval,Augenstein2016SemEval}. Moreover, our approach takes into account the interaction between users on social media, whether it is about appealing for more information in order to corroborate a rumourous statement (\textit{querying}) or to post a response that does not contribute to the resolution of the rumour's veracity (\textit{commenting}). Finally -- and importantly -- instead of dealing with tweets as single units in isolation, we exploit the emergent structure of interactions between users on Twitter, building a classifier that learns the dynamics of stance in tree-structured conversational threads by exploiting its underlying interactional features. While these interactional features do not, in the final analysis, map directly onto those of conversation as revealed by Conversation Analysis \cite{sacks1974simplest}, we argue that there are sufficient relational similarities to justify this approach \cite{tolmie2017ugc}. The closest work is by Ritter et al. \cite{ritter2010unsupervised} who modelled linear sequences of replies in Twitter conversational threads with Hidden Markov Models for dialogue act tagging, but the tree structure of the thread as a whole was not exploited.

As we were writing this article, we also organised, in parallel, a shared task on rumour stance classification, RumourEval \cite{derczynski2017semeval}, at the well-known natural language processing competition SemEval 2017. The subtask A consisted in stance classification of individual tweets discussing a rumour within a conversational thread as one of \textit{support}, \textit{deny}, \textit{query}, or \textit{comment}, which specifically addressed the task presented in this paper. Eight participants submitted results to this task, including work by \cite{kochkina2017turing} using an LSTM classifier which is being also analysed in this paper. In this shared task, most of the systems viewed this task as a 4-way single tweet classification task, with the exception of the best performing system by \cite{kochkina2017turing}, as well as the systems by \cite{wang2017ecnu} and \cite{singh2017iitp}. The winning system addressed the task as a sequential classification problem, where the stance of each tweet takes into consideration the features and labels of the previous tweets. The system by Singh et al. \cite{singh2017iitp} takes as input pairs of source and reply tweets, whereas Wang et al. \cite{wang2017ecnu} addressed class imbalance by decomposing the problem into a two step classification task, first distinguishing between comments and non-comments, to then classify non-comment tweets as one of support, deny or query. Half of the systems employed ensemble classifiers, where classification was obtained through majority voting \cite{wang2017ecnu,garcialozano2017mama,bahuleyan2017uwaterloo,srivastava2017dfki}. In some cases the ensembles were hybrid, consisting both of machine learning classifiers and manually created rules with differential weighting of classifiers for different class labels \cite{wang2017ecnu,garcialozano2017mama,srivastava2017dfki}. Three systems used deep learning, with \cite{kochkina2017turing} employing LSTMs for sequential classification, Chen et al. \cite{chen2017ikm} used convolutional neural networks (CNN) for obtaining the representation of each tweet, assigned a probability for a class by a softmax classifier and Garc\'ia Lozano et al. \cite{garcialozano2017mama} used CNN as one of the classifiers in their hybrid conglomeration. The remaining two systems by Enayet et al. \cite{enayet2017niletmrg} and Singh et al. \cite{singh2017iitp} used support vector machines with a linear and polynomial kernel respectively. Half of the systems invested in elaborate feature engineering, including cue words and expressions denoting Belief, Knowledge, Doubt and Denial \cite{bahuleyan2017uwaterloo} as well as Tweet domain features, including meta-data about users, hashtags and event specific keywords \cite{wang2017ecnu,bahuleyan2017uwaterloo,singh2017iitp,enayet2017niletmrg}. The systems with the least elaborate features were Chen et al. \cite{chen2017ikm} and Garc\'ia Lozano et al. \cite{garcialozano2017mama} for CNNs (word embeddings), Srivastava et al. \cite{srivastava2017dfki} (sparse word vectors as input to logistic regression) and Kochkina et al. \cite{kochkina2017turing} (average word vectors, punctuation, similarity between word vectors in current tweet, source tweet and previous tweet, presence of negation, picture, URL). Five out of the eight systems used pre-trained word embeddings, mostly Google News word2vec embeddings\footnote{\url{https://github.com/mmihaltz/word2vec-GoogleNews-vectors}}, whereas \cite{wang2017ecnu} used four different types of embeddings. The winning system used a sequential classifier, however the rest of the participants opted for other alternatives.

To the best of our knowledge Twitter conversational thread structure has not been explored in detail in the stance classification problem. Here we extend the experimentation presented in our previous work using Conditional Random Fields for rumour stance classification \cite{zubiaga2016coling} in a number of directions: (1) we perform a comparison of a broader range of classifiers, including state-of-the-art rumour stance classifiers such as Hawkes Processes introduced by Lukasik et al. \cite{lukasik2016hawkes}, as well as a new LSTM classifier, (2) we analyse the utility of a larger set of features, including not only local features as in our previous work, but also contextual features that further model the conversational structure of Twitter threads, (3) we perform a more exhaustive analysis of the results, and (4) we perform an analysis of features that gives insight into what characterises the different kinds of stances observed around rumours in social media.

\section{Research Objectives}

The main objective of our research is to analyse whether, the extent to which and how the sequential structure of social media conversations can be exploited to improve the classification of the stance expressed by different posts towards the topic under discussion. Each post in a conversation makes its own contribution to the discussion and hence has to be assigned its own stance value. However, posts in a conversation contribute to previous posts, adding up to a discussion attempting to reach a consensus. Our work looks into the exploitation of this evolving nature of social media discussions with the aim of improving the performance of a stance classifier that has to determine the stance of each tweet. We set forth the following six research objectives:

\textbf{RO 1.} \textit{Quantify performance gains of using sequential classifiers compared with the use of non-sequential classifiers.}

Our first research objective aims to analyse how the use of a sequential classifier that models the evolving nature of social media conversations can perform better than standard classifiers that treat each post in isolation. We do this by solely using local features to represent each post, so that the analysis focuses on the benefits of the sequential classifiers.

\textbf{RO 2.} \textit{Quantify the performance gains using contextual features extracted from the conversation.}

With our second research objective we are interested in analysing whether the use of contextual features (i.e. using other tweets surrounding in a conversation to extract the features of a given tweet) are helpful to boost the classification performance. This is particularly interesting in the case of tweets as they are very short, and inclusion of features extracted from surrounding tweets would be especially helpful. The use of contextual features is motivated by the fact that tweets in a discussion are adding to each other, and hence they cannot be treated alone.

\textbf{RO 3.} \textit{Evaluate the consistency of classifiers across different datasets.}

Our aim is to build a stance classifier that will generalise to multiple different datasets comprising data belonging to different events. To achieve this, we evaluate our classifiers on eight different events.

\textbf{RO 4.} \textit{Assess the effect of the depth of a post in its classification performance.}

We want to build a classifier that will be able to classify stances of different posts occurring at different levels of depth in a conversation. A post can be from a source tweet that initiates a conversation, to a nested reply that occurs later in the sequence formed by a conversational thread. The difficulty increases as replies are deeper as there is more preceding conversation to be aggregated for the classification task. We assess the performance over different depths to evaluate this.

\textbf{RO 5.} \textit{Perform an error analysis to assess when and why each classifier performs best.}

We want to look at the errors made by each of the classifiers. This will help us understand when we are doing well and why, as well as in what cases and with which types of labels we need to keep improving.

\textbf{RO 6.} \textit{Perform an analysis of features to understand and characterise stances in social media discussions.}

In our final objective we are interested in performing an exploration of different features under study, which is informative in two different ways. On the one hand, to find out which features are best for a stance classifier and hence improve performance; on the other hand, to help characterise the different types of stances and hence further understand how people respond in social media discussions.

\section{Rumour Stance Classification}

In what follows we formally define the rumour stance classification task, as well as the datasets we use for our experiments.

\subsection{Task Definition}

The rumour stance classification task consists in determining the type of orientation that each individual post expresses towards the disputed veracity of a rumour. We define the rumour stance classification task as follows: we have a set of conversational threads, each discussing a rumour, $D = \{C_1, ..., C_n\}$. Each conversational thread $C_j$ has a variably sized set of tweets $|C_j|$ discussing it, with a source tweet (the root of the tree), $t_{j,1}$, that initiates it. The source tweet $t_{j,1}$ can receive replies by a varying number $k$ of tweets $Replies_{t_{j,1}} = \{t_{j,1,1}, ..., t_{j,1,k}\}$, which can in turn receive replies by a varying number $k$ of tweets, e.g., $Replies_{t_{j,1,1}} = \{t_{j,1,1,1}, ..., t_{j,1,1,k}\}$, and so on. An example of a conversational thread is shown in Figure \ref{fig:example}.

The task consists in determining the stance of each of the tweets $t_j$ as one of $Y = \{supporting, denying, querying, commenting\}$.

\subsection{Dataset}

As part of the PHEME project \cite{derczynski2015pheme}, we collected a rumour dataset associated with eight events corresponding to breaking news events \cite{zubiaga2016analysing}.\footnote{The entire dataset included nine events, but here we describe the eight events with tweets in English, which we use for our classification experiments. The ninth dataset with tweets in German was not considered for this work.} Tweets in this dataset include tree-structured conversations, which are initiated by a tweet about a rumour (source tweet) and nested replies that further discuss the rumour circulated by the source tweet (replying tweets). The process of collecting the tree-structured conversations initiated by rumours, i.e. having a rumour discussed in the source tweet, and associated with the breaking news events under study was conducted with the assistance of journalist members of the Pheme project team. Tweets comprising the rumourous tree-structured conversations were then annotated for stance using CrowdFlower\footnote{https://www.crowdflower.com/} as a crowdsourcing platform. The annotation process is further detailed in \cite{zubiaga2015crowdsourcing}.

The resulting dataset includes 4,519 tweets and the transformations of annotations described above only affect 24 tweets (0.53\%), i.e., those where the source tweet denies a rumour, which is rare. The example in Figure \ref{fig:example} shows a rumour thread taken from the dataset along with our inferred annotations, as well as how we establish the depth value of each tweet in the thread.

\begin{figure*}
 \footnotesize
 \begin{framed}
  \textit{[depth=0]} \noindent \textbf{u1:} These are not timid colours; soldiers back guarding Tomb of Unknown Soldier after today's shooting \#StandforCanada --PICTURE-- \textbf{[support]}
  \begin{addmargin}[2em]{0pt}
   \textit{[depth=1]} \textbf{u2:} @u1 Apparently a hoax. Best to take Tweet down. \textbf{[deny]}
  \end{addmargin}
  \begin{addmargin}[2em]{0pt}
   \textit{[depth=1]} \textbf{u3:} @u1 This photo was taken this morning, before the shooting. \textbf{[deny]}
  \end{addmargin}
  \begin{addmargin}[2em]{0pt}
   \textit{[depth=1]} \textbf{u4:} @u1 I don't believe there are soldiers guarding this area right now. \textbf{[deny]}
  \end{addmargin}
  \begin{addmargin}[4em]{0pt}
   \textit{[depth=2]} \textbf{u5:} @u4 wondered as well. I've reached out to someone who would know just to confirm that. Hopefully get response soon. \textbf{[comment]}
  \end{addmargin}
  \begin{addmargin}[6em]{0pt}
   \textit{[depth=3]} \textbf{u4:} @u5 ok, thanks. \textbf{[comment]}
  \end{addmargin}
 \end{framed}
 \caption{Example of a tree-structured thread discussing the veracity of a rumour, where the label associated with each tweet is the target of the rumour stance classification task.}
 \label{fig:example}
\end{figure*}

One important characteristic of the dataset, which affects the rumour stance classification task, is that the distribution of categories is clearly skewed towards \textit{commenting} tweets, which account for over 64\% of the tweets. This imbalance varies slightly across the eight events in the dataset (see Table \ref{tab:dataset-stats}). Given that we consider each event as a separate fold that is left out for testing, this varying imbalance makes the task more realistic and challenging. The striking imbalance towards \textit{commenting} tweets is also indicative of the increased difficulty with respect to previous work on stance classification, most of which performed binary classification of tweets as supporting or denying, which account for less than 28\% of the tweets in our case representing a real world scenario.

\begin{table}[htb]
 \footnotesize
 \centering
 \begin{tabular}{ l c c c c c }
  \toprule
  Event & Supporting & Denying & Querying & Commenting & Total \\
  \midrule
  charliehebdo & 239 (22.0\%) & 58 (5.0\%) & 53 (4.0\%) & 721 (67.0\%) & 1,071 \\
  ebola-essien & 6 (17.0\%) & 6 (17.0\%) & 1 (2.0\%) & 21 (61.0\%) & 34 \\
  ferguson & 176 (16.0\%) & 91 (8.0\%) & 99 (9.0\%) & 718 (66.0\%) & 1,084 \\
  germanwings-crash & 69 (24.0\%) & 11 (3.0\%) & 28 (9.0\%) & 173 (61.0\%) & 281 \\
  ottawashooting & 161 (20.0\%) & 76 (9.0\%) & 63 (8.0\%) & 477 (61.0\%) & 777 \\
  prince-toronto & 21 (20.0\%) & 7 (6.0\%) & 11 (10.0\%) & 64 (62.0\%) & 103 \\
  putinmissing & 18 (29.0\%) & 6 (9.0\%) & 5 (8.0\%) & 33 (53.0\%) & 62 \\
  sydneysiege & 220 (19.0\%) & 89 (8.0\%) & 98 (8.0\%) & 700 (63.0\%) & 1,107 \\
  \midrule
  Total & 910 (20.1\%) & 344 (7.6\%) & 358 (7.9\%) & 2,907 (64.3\%) & 4,519 \\
  \bottomrule
 \end{tabular}
 \caption{Distribution of categories for the eight events in the dataset.}
 \label{tab:dataset-stats}
\end{table}

\section{Classifiers}

In this section we describe the different classifiers that we used for our experiments. Our focus is on sequential classifiers, especially looking at classifiers that exploit the discursive nature of social media, which is the case for Conditional Random Fields in two different settings -- i.e. Linear CRF and tree CRF -- as well as that of a Long Short-Term Memory (LSTM) in a linear setting -- Branch LSTM. We also experiment with a sequential classifier based on Hawkes Processes that instead exploits the temporal sequence of tweets and has been shown to achieve state-of-the-art performance \cite{lukasik2016hawkes}. After describing these three types of classifiers, we outline a set of baseline classifiers.

\subsection{Hawkes Processes}

One approach for modelling arrival of tweets around rumours is based on point processes, a probabilistic framework where tweet occurrence likelihood is modelled using an intensity function over time. Intuitively, higher values of intensity function denote higher likelihood of tweet occurrence. For example, Lukasik et al. modelled tweet occurrences over time with a log-Gaussian Cox Process, a point process which models its intensity function as an exponentiated sample of a Gaussian Process \cite{lukasik15_dynamics,lukasik15_tweetarrival,lukasik16_conv}. In related work, tweet arrivals were modelled with a Hawkes Process and a resulting model was applied for stance classification of tweets around rumours \cite{lukasik2016hawkes}. In this subsection we describe the sequence classification algorithm based on Hawkes Processes.

\paragraph{Intensity Function}
The intensity function in a Hawkes Process is expressed as a summation of base intensity and the intensities corresponding to influences of previous tweets,
\begin{flalign}
\label{eq:HawkesCI2}
\lambda_{y,m}(t)\!\! =\!\! \displaystyle \mu_y \! +\!\! \sum_{t_\ell < t} \mathbb{I}(m_\ell = m)
\alpha_{y_\ell,y} \kappa(t - t_\ell),
\end{flalign}
where the first term  represents the constant base intensity of generating label $y$. The second term represents the influence from the previous tweets. The influence from each tweet is modelled with an exponential kernel function $\kappa(t - t_\ell) = \omega \exp(-\omega (t - t_\ell))$. 
The matrix $\alpha$ of size $|Y| \times |Y|$ encodes how pairs of labels corresponding to tweets influence one another, e.g. how a \emph{querying} label influences a \emph{rejecting} label.

\paragraph{Likelihood function}
The parameters governing the intensity function are learnt by maximising the likelihood of generating the tweets:

\begin{flalign}
\label{eq:factorizedCL}
L(\bm{t}, \bm{y}, \bm{m}, \bm{W})  = \prod_{n=1}^N  p(\bm{W}_n | y_n)  \times \Big [ \prod_{n=1}^N \lambda_{y_n,m_n}(t_n) \Big ] \! \times \! p(E_T),
\end{flalign}

where the likelihood of generating text given the label is modelled as a multinomial distribution conditioned on the label (parametrised by matrix $\beta$). The second term provides the likelihood of occurrence of tweets at times $t_1, \ldots , t_n$ and the third term provides the likelihood that no tweets happen in the interval $[0,T]$ except at times $t_1, \ldots, t_n$. We estimate the parameters of the model by maximising the log-likelihood. As in \cite{lukasik2016hawkes}, Laplacian smoothing is applied to the estimated language parameter $\beta$.

In one approach to $\mu$ and $\alpha$ optimisation (\textit{Hawkes Process with Approximated Likelihood}, or \emph{HP Approx.} \cite{lukasik2016hawkes}) a closed form updates for $\mu$ and $\alpha$ are obtained using an approximation of the log-likelihood of the data. In a different approach (\textit{Hawkes Process with Exact Likelihood}, or \emph{HP Grad.} \cite{lukasik2016hawkes}) parameters are found using joint gradient based optimisation over $\mu$ and $\alpha$, using derivatives of log-likelihood\footnote{For both implementations we used the `seqhawkes' Python package: \url{https://github.com/mlukasik/seqhawkes}}. L-BFGS approach is employed for gradient search. Parameters are initialised with those found by the \emph{HP Approx.} method. Moreover, following previous work we fix the decay parameter $\omega$ to $0.1$.

We predict the most likely sequence of labels, thus maximising the likelihood of occurrence of the tweets from Equation (\ref{eq:factorizedCL}), or the approximated likelihood in case of \emph{HP Approx.} Similarly as in \cite{lukasik2016hawkes}, we follow a greedy approach, where we choose the most likely label for each consecutive tweet. 

\subsection{Conditional Random Fields (CRF): Linear CRF and Tree CRF}

We use CRF as a structured classifier to model sequences observed in Twitter conversations. With CRF, we can model the conversation as a graph that will be treated as a sequence of stances, which also enables us to assess the utility of harnessing the conversational structure for stance classification. Different to traditionally used classifiers for this task, which choose a label for each input unit (e.g. a tweet), CRF also consider the neighbours of each unit, learning the probabilities of transitions of label pairs to be followed by each other. The input for CRF is a graph $G = (V, E)$, where in our case each of the vertices $V$ is a tweet, and the edges $E$ are relations of tweets replying to each other. Hence, having a data sequence $X$ as input, CRF outputs a sequence of labels $Y$ \cite{lafferty2001conditional}, where the output of each element $y_i$ will not only depend on its features, but also on the probabilities of other labels surrounding it. The generalisable conditional distribution of CRF is shown in Equation \ref{eq:crf} \cite{sutton2011introduction}.

\begin{equation}
 p(y|x) = \frac{1}{Z(x)} \prod_{a = 1}^{A} \Psi_a (y_a, x_a)
 \label{eq:crf}
\end{equation}

where Z(x) is the normalisation constant, and $\Psi_a$ is the set of factors in the graph $G$.

\begin{figure*}[ht]
 \centering
 \includegraphics[width=0.8\textwidth]{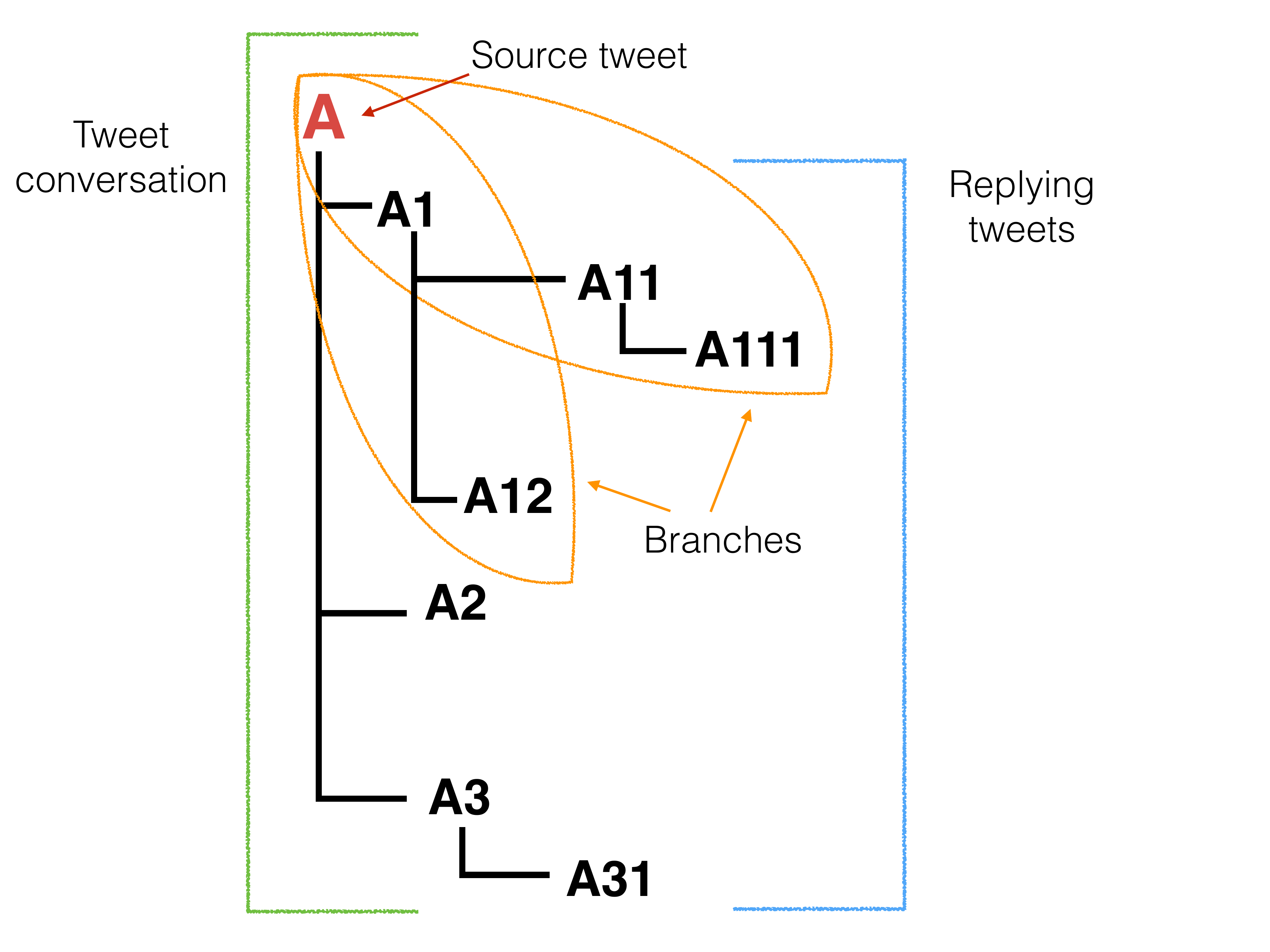}
 \caption{Example of a tree-structured conversation, with two overlapping branches highlighted.}
 \label{fig:tree-and-branches}
\end{figure*}

We use CRFs in two different settings.\footnote{We use the PyStruct to implement both variants of CRF \cite{muller2014pystruct}.} First, we use a linear-chain CRF (Linear CRF) to model each branch as a sequence to be input to the classifier. We also use Tree-Structured CRFs (Tree CRF) or General CRFs to model the whole, tree-structured conversation as the sequence input to the classifier. So in the first case the sequence unit is a branch and our input is a collection of branches and in the second case our sequence unit is an entire conversation, and our input is a collection of trees. An example of the distinction of dealing with branches or trees is shown in Figure \ref{fig:tree-and-branches}. With this distinction we also want to experiment whether it is worthwhile building the whole tree as a more complex graph, given that users replying in one branch might not have necessarily seen and be conditioned by tweets in other branches. However, we believe that the tendency of types of replies observed in a branch might also be indicative of the distribution of types of replies in other branches, and hence useful to boost the performance of the classifier when using the tree as a whole. An important caveat of modelling a tree in branches is also that there is a need to repeat parts of the tree across branches, e.g., the source tweet will repeatedly occur as the first tweet in every branch extracted from a tree.\footnote{Despite this also leading to having tweets repeated across branches in the test set and hence producing an output repeatedly for the same tweet with Linear CRF, this output does is consistent and there is no need to aggregate different outputs.}

To account for the imbalance of classes in our datasets, we perform cost-sensitive learning by assigning weighted probabilities to each of the classes, these probabilities being the inverse of the number of occurrences observed in the training data for a class.

\subsection{Branch LSTM}

\begin{figure*}[ht]
 \centering
 \includegraphics[width=0.7\textwidth]{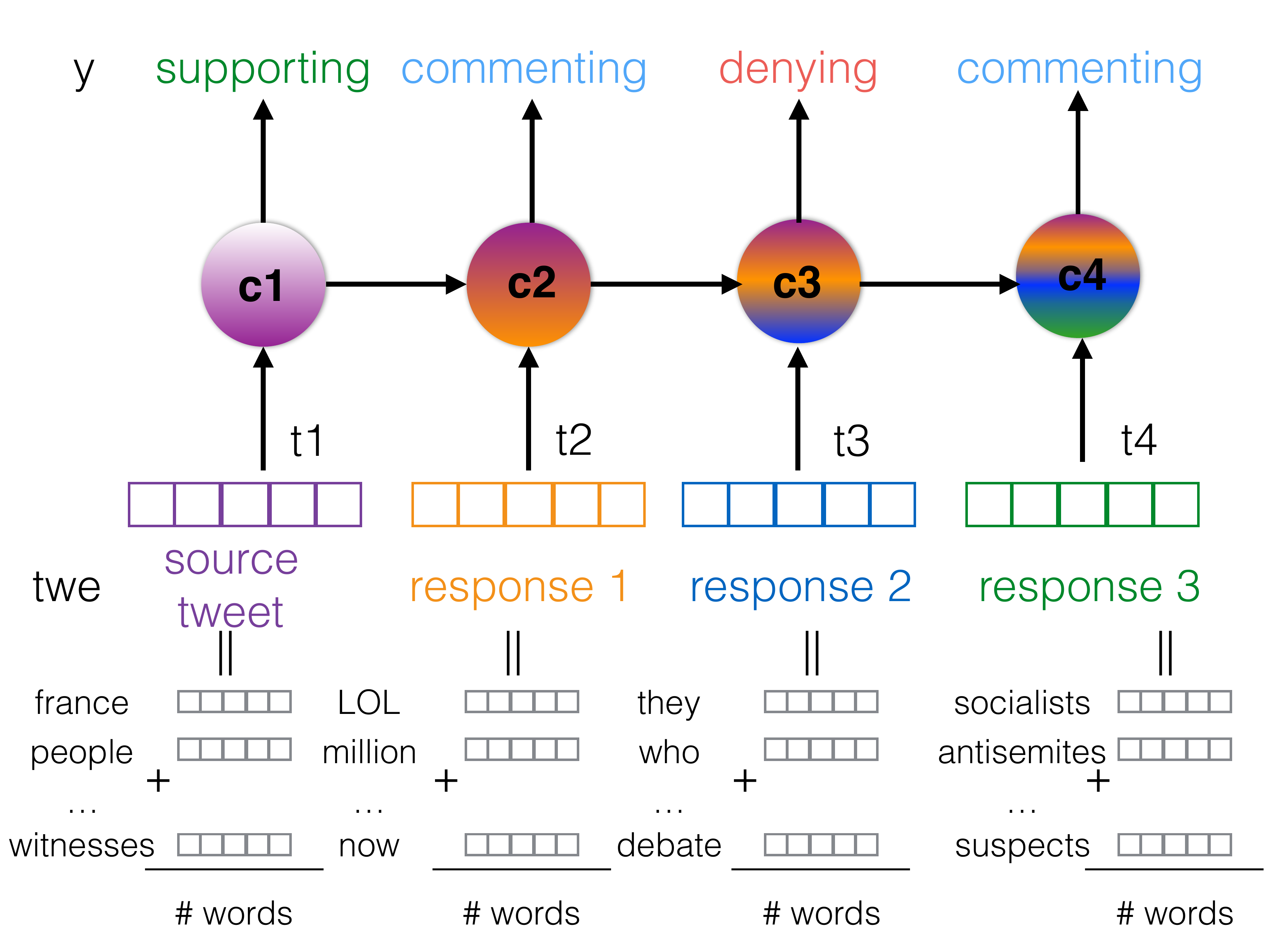}
 \caption{Illustration of the input/output structure of the LSTM-branch model}
 \label{fig:LSTMbranchio}
\end{figure*}

Another model that works with structured input is a neural network with Long Short-Term Memory (LSTM) units \cite{hochreiter1997long}. LSTMs are able to model discrete time series and possess a `memory' property of the previous time steps, therefore we propose a \textit{branch-LSTM} model that utilises them to process branches of tweets.

Figure \ref{fig:LSTMbranchio} illustrates how the input of the time step of the LSTM layer is a vector that is an average of word vectors from each tweet and how the information propagates between time steps. 

The full model consists of several LSTM layers that are connected to several feed-forward ReLU layers and a softmax layer to obtain predicted probabilities of a tweet belonging to certain class. As a means for weight regularisation we utilise \textit{dropout} and \textit{l2-norm}. We use categorical cross-entropy as the loss function. The model is trained using mini-batches and the Adam optimisation algorithm \cite{kingma2014adam}.\footnote{For implementation of all models we used Python libraries Theano \cite{bastien2012theano} and Lasagne \cite{lasagne}.}

The number of layers, number of units in each layer, regularisation strength, mini-batch size and learning rate are determined using the Tree of Parzen Estimators (TPE) algorithm \cite{bergstra2011algorithms}\footnote{We use the implementation in the hyperopt package \cite{bergstra2013making}.} on the development set.\footnote{For this setting, we use the `Ottawa shooting' event for development.} 

The \textit{branch-LSTM} takes as input tweets represented as the average of its word vectors. We also experimented with obtaining tweet representations through per-word nested LSTM layers, however, this approach did not result in significantly better results than the average of word vectors.

Extracting branches from a tree-structured conversation presents the caveat that some tweets are repeated across branches after this conversion. We solve this issue by applying a mask to the loss function to not take repeated tweets into account.

\subsection{Summary of Sequential Classifiers}

All of the classifiers described above make use of the sequential nature of Twitter conversational threads. These classifiers take a sequence of tweets as input, where the relations between tweets are formed by replies. If C replies to B, and B replies to A, it will lead to a sequence ``A $\rightarrow$ B $\rightarrow$ C''. Sequential classifiers will use the predictions on preceding tweets to determine the possible label for each tweet. For instance, the classification for B will depend on the prediction that has been previously made for A, and the probabilities of different labels for B will vary for the classifier depending on what has been predicted for A.

Among the four classifiers described above, the one that differs in how the sequence is treated is the Tree CRF. This classifier builds a tree-structured graph with the sequential relationships composed by replying tweets. The rest of the classifiers, Hawkes Processes, Linear CRF and LSTM, will break the entire conversational tree into linear branches, and the input to the classifiers will be linear sequences. The use of a graph with the Tree CRF has the advantage of building a single structure, while the rest of the classifiers building linear sequences inevitably need to repeat tweets across different linear sequences. All the linear sequences will repeatedly start with the source tweet, while some of the subsequent tweets may also be repeated. The use of linear sequences has however the advantages of simplifying the model being used, and one may also hypothesise that inclusion of the entire tree made of different branches into the same graph may not be suitable when they may all be discussing issues that differ to some extent from one another. Figure \ref{fig:tree-and-branches} shows an example of a conversation tree, how the entire tree would make a graph, as well as how we break it down into smaller branches or linear sequences.

\subsection{Baseline Classifiers}

\noindent \textbf{Maximum Entropy classifier (MaxEnt).} As the non-sequential counterpart of CRF, we use a Maximum Entropy (or logistic regression) classifier, which is also a conditional classifier but which will operate at the tweet level, ignoring the conversational structure. This enables us to directly compare the extent to which treating conversations as sequences instead of having each tweet as a separate unit can boost the performance of the CRF classifiers. We perform cost-sensitive learning by assigning weighted probabilities to each class as the inverse of the number of occurrences in the training data.

\noindent \textbf{Additional baselines.} We also compare two more non-sequential classifiers\footnote{We use their implementation in the scikit-learn Python package, using the \textit{class\_weight=``balanced''} parameter to perform cost-sensitive learning.}: Support Vector Machines (SVM), and Random Forests (RF).

\subsection{Experiment Settings and Evaluation Measures}

We experiment in an 8-fold cross-validation setting. Given that we have 8 different events in our dataset, we create 8 different folds, each having the data linked to an event. In our cross-validation setting, we run the classifier 8 times, on each occasion having a different fold for testing, with the other 7 for training. In this way, each fold is tested once, and the aggregation of all folds enables experimentation on all events. For each of the events in the test set, the experiments consist in classifying the stance of each individual tweet. With this, we simulate a realistic scenario where we need to use knowledge from past events to train a model that will be used to classify tweets in new events.

Given that the classes are clearly imbalanced in our case, evaluation based on accuracy arguably cannot suffice to capture competitive performance beyond the majority class. To account for the imbalance of the categories, we report the macro-averaged F1 scores, which measures the overall performance assigning the same weight to each category. We aggregate the macro-averaged F1 scores to get the final performance score of a classifier. We also use the McNemar test \cite{mcnemar1947note} throughout the analysis of results to further compare the performance of some classifiers.

It is also worth noting that all the sequential classifiers only make use of preceding tweets in the conversation to classify a tweet, and hence no later tweets are used. That is the case of a sequence \textit{$t_1$, $t_2$, $t_3$} of tweets, each responding to the preceding tweet. The sequential classifier attempting to classify $t_2$ would incorporate $t_1$ in the sequence, but $t_3$ would not be considered.

\section{Features}

While focusing on the study of sequential classifiers for discursive stance classification, we perform our experiments with three different types of features: local features, contextual features and Hawkes features. First, local features enable us to evaluate the performance of sequential classifiers in a comparable setting to non-sequential classifiers where features are extracted solely from the current tweet; this makes it a fairer comparison where we can quantify the extent to which mining sequences can boost performance. In a subsequent step, we also incorporate contextual features, i.e. features from other tweets in a conversation, which enables us to further boost performance of the sequential classifiers. Finally, and to enable comparison with the Hawkes process classifier, we describe the Hawkes features.

Table \ref{tab:features} shows the list of features used, both local and contextual, each of which can be categorised into several subtypes of features, as well as the Hawkes features. For more details on these features, please see \ref{ap:features}.

\linespread{1}

\begin{table}[htb]
 \footnotesize
 \centering
 \begin{tabular*}{0.66\textwidth}{l | l}
  \toprule
  \multicolumn{2}{c}{\textbf{Local features}} \\
  \toprule
  \multirow{4}{*}{\textbf{Lexicon}} & Word embeddings \\
   & POS tags \\
   & Negation \\
   & Swear words \\
  \midrule
  \multirow{2}{*}{\textbf{Content formatting}} & Tweet length \\
   & Word count \\
  \midrule
  \multirow{2}{*}{\textbf{Punctuation}} & Question mark \\
   & Exclamation mark \\
  \midrule
  \multirow{1}{*}{\textbf{Tweet formatting}} & URL attached \\
  \toprule
  \multicolumn{2}{c}{\textbf{Contextual features}} \\
  \toprule
  \multirow{3}{*}{\textbf{Relational}} & Word2Vec similarity wrt source tweet \\
   & Word2Vec similarity wrt preceding tweet \\
   & Word2Vec similarity wrt thread \\
  \midrule
  \multirow{4}{*}{\textbf{Structural}} & Is leaf \\
   & Is source tweet \\
   & Is source user \\
  \midrule
  \multirow{3}{*}{\textbf{Social}} & Has favourites \\
   & Has retweets \\
   & Persistence \\
   & Time difference \\
  \toprule
  \multicolumn{2}{c}{\textbf{Hawkes features}} \\
  \toprule
  \toprule
  \multirow{2}{*}{\textbf{Hawkes features}} & Bag of words \\
   & Timestamp \\
  \bottomrule
 \end{tabular*}
 \caption{List of features.}
 \label{tab:features}
\end{table}

\linespread{2}

\section{Experimental Results}
\label{sec:experimental-results}

\subsection{Evaluating Sequential Classifiers (RO 1)}
\label{sec:eval-sequential}

First, we evaluate the performance of the classifiers by using only local features. As noted above, this enables us to perform a fairer comparison of the different classifiers by using features that can be obtained solely from each tweet in isolation; likewise, it enables us to assess whether and the extent to which the use of a sequential classifier to exploit the discursive structure of conversational threads can be of help to boost performance of the stance classifier while using the same set of features as non-sequential classifiers.

Therefore, in this section we make use of the local features described in Section \ref{ssec:local-features}. Additionally, we also use the Hawkes features described in Section \ref{ssec:hawkes-features} for comparison with the Hawkes processes. For the set of local features, we show the results for three different scenarios: (1) using each subgroup of features alone, (2) in a leave-one-out setting where one of the subgroups is not used, and (3) using all of the subgroups combined.

Table \ref{tab:results-sequence} shows the results for the different classifiers using the combinations of local features as well as Hawkes features. We make the following observations from these results:

\begin{itemize}
 \item LSTM consistently performs very well with different features.
 \item Confirming our main hypothesis and objective, sequential classifiers do show an overall superior performance to the non-sequential classifiers. While the two CRF alternatives perform very well, the LSTM classifier is slightly superior (the differences between CRF and LSTM results are statistically significant at $p < 0.05$, except for the LF1 features). Moreover, the CRF classifiers outperform their non-sequential counterpart MaxEnt, which achieves an overall lower performance (all the differences between CRF and MaxEnt results being statistically significant at $p < 0.05$).
 \item The LSTM classifier is, in fact, superior to the Tree CRF classifier (all statistically significant except LF1). While the Tree CRF needs to make use of the entire tree for the classification, the LSTM classifier only uses branches, reducing the amount of data and complexity that needs to be processed in each sequence.
 \item Among the local features, combinations of subgroups of features lead to clear improvements with respect to single subgroups without combinations.
 \item Even though the combination of all local features achieves good performance, there are alternative leave-one-out combinations that perform better. The feature combination leading to the best macro-F1 score is that combining lexicon, content formatting and punctuation (i.e. LF123, achieving a score of 0.449).
\end{itemize}

Summarising, our initial results show that exploiting the sequential properties of conversational threads, while still using only local features to enable comparison, leads to superior performance with respect to the classification of each tweet in isolation by non-sequential classifiers. Moreover, we observe that the local features combining lexicon, content formatting and punctuation lead to the most accurate results. In the next section we further explore the use of contextual features in combination with local features to boost performance of sequential classifiers; to represent the local features, we rely on the best approach from this section (i.e. LF123).

\linespread{1}

\begin{table}[htb]
  \scriptsize
  \centering
  \begin{tabular}{l || l l l l l l l l l l }
   \toprule
   \multicolumn{11}{c}{\textbf{Macro-F1}} \\
   \toprule
    & HF & LF1 & LF2 & LF3 & LF4 & LF123 & LF124 & LF134 & LF234 & LF1234 \\
   \midrule
   SVM & 0.336 & 0.356 & 0.231 & 0.258 & 0.313 & 0.403 & 0.365 & 0.403 & 0.420 & 0.408 \\
   Random Forest & 0.325 & 0.308 & 0.276 & 0.267 & \textbf{0.437*} & 0.322 & 0.310 & 0.351 & 0.357 & 0.329 \\
   MaxEnt & 0.338 & 0.363 & 0.272 & 0.263 & 0.428 & 0.415 & 0.363 & 0.421 & 0.427 & 0.422 \\
   \midrule
   Hawkes-approx & 0.309 & -- & -- & -- & -- & -- & -- & -- & -- & -- \\
   Hawkes-grad & 0.307 & -- & -- & -- & -- & -- & -- & -- & -- & -- \\
   Linear CRF & \textbf{0.362*} & 0.357 & 0.268 & 0.318 & 0.317 & 0.413 & 0.365 & 0.403 & 0.425 & 0.412 \\
   Tree CRF & 0.350 & \textbf{0.375*} & 0.285 & 0.221 & 0.217 & 0.433 & 0.385 & \textbf{0.413} & \textbf{0.436*} & 0.433 \\
   LSTM & 0.318 & 0.362 & \textbf{0.318*} & \textbf{0.407*} & 0.419 & \textbf{0.449*} & \textbf{0.395*} & 0.412 & 0.429 & \textbf{0.437*} \\
   \bottomrule
  \end{tabular}
  \caption{Macro-F1 performance results using local features. HF: Hawkes features. LF: local features, where numbers indicate subgroups of features as follows, 1: Lexicon, 2: Content formatting, 3: Punctuation, 4: Tweet formatting. An '*' indicates that the differences between the best performing classifier and the second best classifier for that feature set are statistically significant at $p < 0.05$.}
  \label{tab:results-sequence}
\end{table}

\linespread{2}

\subsection{Exploring Contextual Features (RO 2)}
\label{sec:eval-contextual}

The experiments in the previous section show that sequential classifiers that model discourse, especially the LSTM classifier, can provide substantial improvements over non-sequential classifiers that classify each tweet in isolation, in both cases using only local features to represent each tweet. To complement this, we now explore the inclusion of contextual features described in Section \ref{ssec:contextual-features} for the stance classification. We perform experiments with four different groups of features in this case, including local features and the three subgroups of contextual features, namely relational features, structural features and social features. As in the previous section, we show results for the use of each subgroup of features alone, in a leave-one-out setting, and using all subgroups of features together.

Table \ref{tab:results-contextual} shows the results for the classifiers incorporating contextual features along with local features. We make the following observations from these results:

\begin{itemize}
 \item The use of contextual features leads to substantial improvements for non-sequential classifiers, getting closer to and even in some cases outperforming some of the sequential classifiers.
 \item Sequential classifiers, however, do not benefit much from using contextual features. It is important to note that sequential classifiers are taking the surrounding context into consideration when they aggregate sequences in the classification process. This shows that the inclusion of contextual features is not needed for sequential classifiers, given that they are implicitly including context through the use of sequences.
 \item In fact, for the LSTM, which is still the best-performing classifier, it is better to only rely on local features, as the rest of the features do not lead to any improvements. Again, the LSTM is able to handle context on its own, and therefore inclusion of contextual features is redundant and may be harmful.
 \item Addition of contextual features leads to substantial improvements for the non-sequential classifiers, achieving similar macro-averaged scores in some cases (e.g. MaxEnt / All vs LSTM / LF). This reinforces the importance of incorporating context in the classification process, which leads to improvements for the non-sequential classifier when contextual features are added, but especially in the case of sequential classifiers that can natively handle context.
\end{itemize}

\linespread{1}

\begin{table}[htb]
  \scriptsize
  \centering
  \begin{tabular}{l || l l l l l l l l l }
   \toprule
   \multicolumn{10}{c}{\textbf{Macro-F1}} \\
   \toprule
    & LF & R & ST & SO & LF+R+ST & LF+R+SO & LF+ST+SO & R+ST+SO & All \\
   \midrule
   SVM & 0.403 & \textbf{0.335*} & \textbf{0.318} & 0.260 & 0.429 & 0.347 & 0.388 & 0.295 & 0.375 \\
   Random Forest & 0.322 & 0.325 & 0.269 & 0.328 & 0.356 & 0.358 & 0.376 & \textbf{0.343*} & 0.364 \\
   MaxEnt & 0.415 & 0.333 & \textbf{0.318} & 0.310 & 0.434 & \textbf{0.447} & 0.447 & 0.318 & \textbf{0.449} \\
   \midrule
   Linear CRF & 0.413 & 0.318 & \textbf{0.318} & \textbf{0.334*} & 0.424 & 0.431 & 0.431 & 0.342 & 0.437 \\
   Tree CRF & 0.433 & 0.322 & 0.317 & 0.312 & 0.425 & 0.429 & 0.430 & 0.232 & 0.433 \\
   LSTM & \textbf{0.449*} & 0.318 & \textbf{0.318} & 0.315 & \textbf{0.445*} & 0.436 & \textbf{0.448} & 0.314 & 0.437 \\
   \bottomrule
  \end{tabular}
  \caption{Macro-F1 performance results incorporating contextual features. LF: local features, R: relational features, ST: structural features, SO: social features. An '*' indicates that the differences between the best performing classifier and the second best classifier for that feature set are statistically significant.}
  \label{tab:results-contextual}
\end{table}

\linespread{2}

Summarising, we observe that the addition of contextual features is clearly useful for non-sequential classifiers, which do not consider context natively. For the sequential classifiers, which natively consider context in the classification process, the inclusion of contextual features is not helpful and is even harmful in most cases, potentially owing to the contextual information being used twice. Still, sequential classifiers, and especially LSTM, are the best classifiers to achieve optimal results, which also avoid the need for computing contextual features.

\subsection{Analysis of the Best-Performing Classifiers}

Despite the clear superiority of LSTM with the sole use of local features, we now further examine the results of the best-performing classifiers to understand when they perform well. We compare the performance of the following five classifiers in this section: (1) LSTM with only local features, (2) Tree CRF with all the features, (3) Linear CRF with all the features, (4) MaxEnt with all the features, and (5) SVM using local features, relational and structural features. Note that while for LSTM we only need local features, for the rest of the classifiers we need to rely on all or almost all of the features. For these best-performing combinations of classifiers and features, we perform additional analyses by event and by tweet depth, and perform an analysis of errors.

\subsubsection{Evaluation by Event (RO 3)}

The analysis of the best-performing classifiers, broken down by event, is shown in Table \ref{tab:results-events}. These results suggest that there is not a single classifier that performs best in all cases; this is most likely due to the diversity of events. However, we see that the LSTM is the classifier that outperforms the rest in the greater number of cases; this is true for three out of the eight cases (the difference with respect to the second best classifier being always statistically significant). Moreover, sequential classifiers perform best in the majority of the cases, with only three cases where a non-sequential classifier performs best. Most importantly, these results suggest that sequential classifiers outperform non-sequential classifiers across the different events under study, with LSTM standing out as a classifier that performs best in numerous cases using only local features.

\linespread{1}

\begin{table}[htb]
  \scriptsize
  \centering
  \begin{tabular}{l || l l l l l l l l }
   \toprule
   \multicolumn{9}{c}{\textbf{Macro-F1}} \\
   \toprule
   & CH & Ebola & Ferg. & GW crash & Ottawa & Prince & Putin & Sydney \\
   \midrule
   SVM & 0.399 & 0.380 & 0.382 & 0.427 & \textbf{0.492} & 0.491 & 0.509 & 0.427 \\
   MaxEnt & 0.446 & 0.425 & \textbf{0.418} & 0.475 & 0.468 & \textbf{0.514} & 0.381 & 0.443 \\
   \midrule
   Linear CRF & 0.443 & 0.619 & 0.380 & 0.470 & 0.412 & 0.512 & \textbf{0.528} & \textbf{0.454} \\
   Tree CRF & 0.457 & 0.557 & 0.356 & 0.523 & 0.441 & 0.505 & 0.491 & 0.426 \\
   LSTM & \textbf{0.465} & \textbf{0.657} & 0.373 & \textbf{0.543} & 0.475 & 0.379 & 0.457 & 0.446 \\
   \bottomrule
  \end{tabular}
  \caption{Macro-F1 results for the best-performing classifiers, broken down by event.}
  \label{tab:results-events}
\end{table}

\linespread{2}

\subsubsection{Evaluation by Tweet Depth (RO 4)}

The analysis of the best-performing classifiers, broken down by depth of tweets, is shown in Table \ref{tab:results-depth}. Note that the depth of the tweet reflects, as shown in Figure \ref{fig:example}, the number of steps from the source tweet to the current tweet. We show results for all the depths from 0 to 4, as well as for the subsequent depths aggregated as 5+.

Again, we see that there is not a single classifier that performs best for all depths. We see, however, that sequential classifiers (Linear CRF, Tree CRF and LSTM) outperform non-sequential classifiers (SVM and MaxEnt) consistently. However, the best sequential classifier varies. While LSTM is the best-performing classifier overall when we look at macro-averaged F1 scores, as shown in Section \ref{sec:eval-contextual}, surprisingly it does not achieve the highest macro-averaged F1 scores at any depth. It does, however, perform well for each depth compared to the rest of the classifiers, generally being close to the best classifier in that case. Its consistently good performance across different depths makes it the best overall classifier, despite only using local features.

\linespread{1}

\begin{table}[htb]
  \scriptsize
  \centering
  \begin{tabular}{l || l l l l l l }
   \toprule
   \multicolumn{7}{c}{\textbf{Tweets by depth}} \\
   \toprule
   & 0 & 1 & 2 & 3 & 4 & 5+ \\
   \midrule
   Counts & 297 & 2,602 & 553 & 313 & 195 & 595 \\
   \toprule
   \multicolumn{7}{c}{\textbf{Macro-F1}} \\
   \toprule
   & 0 & 1 & 2 & 3 & 4 & 5+ \\
   \midrule
   SVM & 0.272 & 0.368 & 0.298 & 0.314 & 0.331 & 0.274 \\
   MaxEnt & 0.238 & 0.385 & 0.286 & 0.279 & \textbf{0.369} & \textbf{0.290} \\
   \midrule
   Linear CRF & \textbf{0.286} & 0.394 & \textbf{0.306} & 0.282 & 0.271 & 0.266 \\
   Tree CRF & 0.278 & \textbf{0.404} & 0.280 & \textbf{0.331} & 0.230 & 0.237 \\
   LSTM & 0.271 & 0.381 & 0.298 & 0.274 & 0.307 & 0.286 \\
   \bottomrule
  \end{tabular}
  \caption{Macro-F1 results for the best-performing classifiers, broken down by tweet depth.}
  \label{tab:results-depth}
\end{table}

\linespread{2}

\subsubsection{Error Analysis (RO 5)}

To analyse the errors that the different classifiers are making, we look at the confusion matrices in Table \ref{tab:confusion}. If we look at the correct guesses, highlighted in bold in the diagonals, we see that the LSTM clearly performs best for three of the categories, namely \textit{support}, \textit{deny} and \textit{query}, and it is just slightly behind the other classifiers for the majority class, \textit{comment}. Besides LSTM's overall superior performance as we observed above, this also confirms that the LSTM is doing better than the rest of the classifiers in dealing with the imbalance inherent in our datasets. For instance, the \textit{Deny} category proves especially challenging for being less common than the rest (only 7.6\% of instances in our datasets); the LSTM still achieves the highest performance for this category, which, however, only achieves 0.212 in accuracy and may benefit from having more training instances.

We also notice that a large number of instances are misclassified as \textit{comments}, due to this being the prevailing category and hence having a much larger number of training instances. One could think of balancing the training instances to reduce the prevalence of \textit{comments} in the training set, however, this is not straightforward for sequential classifiers as one needs to then break sequences, losing not only some instances of \textit{comments}, but also connections between instances of other categories that belong to those sequences. Other solutions, such as labelling more data or using more sophisticated features to distinguish different categories, might be needed to deal with this issue; given that the scope of this paper is to assess whether and the extent to which sequential classifiers can be of help in stance classification, further tackling this imbalance is left for future work.

\linespread{1}

\begin{table}[htb]
  \scriptsize
  \centering
  \begin{tabular}{l || l l l l }
   \toprule
   \multicolumn{5}{c}{\textbf{SVM}} \\
   \toprule
   & Support & Deny & Query & Comment \\
   \midrule
   Support & \textbf{0.657} & 0.041 & 0.018 & 0.283 \\
   Deny & 0.185 & \textbf{0.129} & 0.107 & 0.579 \\
   Query & 0.083 & 0.081 & \textbf{0.343} & 0.494 \\
   Comment & 0.150 & 0.075 & 0.053 & \textbf{0.723} \\
   \toprule
   \multicolumn{5}{c}{\textbf{MaxEnt}} \\
   \toprule
   & Support & Deny & Query & Comment \\
   \midrule
   Support & \textbf{0.794} & 0.044 & 0.003 & 0.159 \\
   Deny & 0.156 & \textbf{0.130} & 0.079 & 0.634 \\
   Query & 0.088 & 0.066 & \textbf{0.366} & 0.480 \\
   Comment & 0.152 & 0.074 & 0.048 & \textbf{0.726} \\
   \toprule
   \multicolumn{5}{c}{\textbf{Linear CRF}} \\
   \toprule
   & Support & Deny & Query & Comment \\
   \midrule
   Support & \textbf{0.603} & 0.048 & 0.013 & 0.335 \\
   Deny & 0.219 & \textbf{0.140} & 0.050 & 0.591 \\
   Query & 0.071 & 0.095 & \textbf{0.357} & 0.476 \\
   Comment & 0.139 & 0.072 & 0.062 & \textbf{0.726} \\
   \toprule
   \multicolumn{5}{c}{\textbf{Tree CRF}} \\
   \toprule
   & Support & Deny & Query & Comment \\
   \midrule
   Support & \textbf{0.552} & 0.066 & 0.019 & 0.363 \\
   Deny & 0.145 & \textbf{0.169} & 0.081 & 0.605 \\
   Query & 0.077 & 0.081 & \textbf{0.401} & 0.441 \\
   Comment & 0.128 & 0.074 & 0.068 & \textbf{0.730} \\
   \toprule
   \multicolumn{5}{c}{\textbf{LSTM}} \\
   \toprule
   & Support & Deny & Query & Comment \\
   \midrule
   Support & \textbf{0.825} & 0.046 & 0.003 & 0.127 \\
   Deny & 0.225 & \textbf{0.212} & 0.125 & 0.438 \\
   Query & 0.090 & 0.087 & \textbf{0.432} & 0.390 \\
   Comment & 0.144 & 0.076 & 0.057 & \textbf{0.723} \\
   \bottomrule
  \end{tabular}
  \caption{Confusion matrices for the best-performing classifiers.}
  \label{tab:confusion}
\end{table}

\linespread{2}

\subsection{Feature Analysis (RO 6)}

\begin{figure}[phtb]
 \centering
 \includegraphics[width=\textwidth]{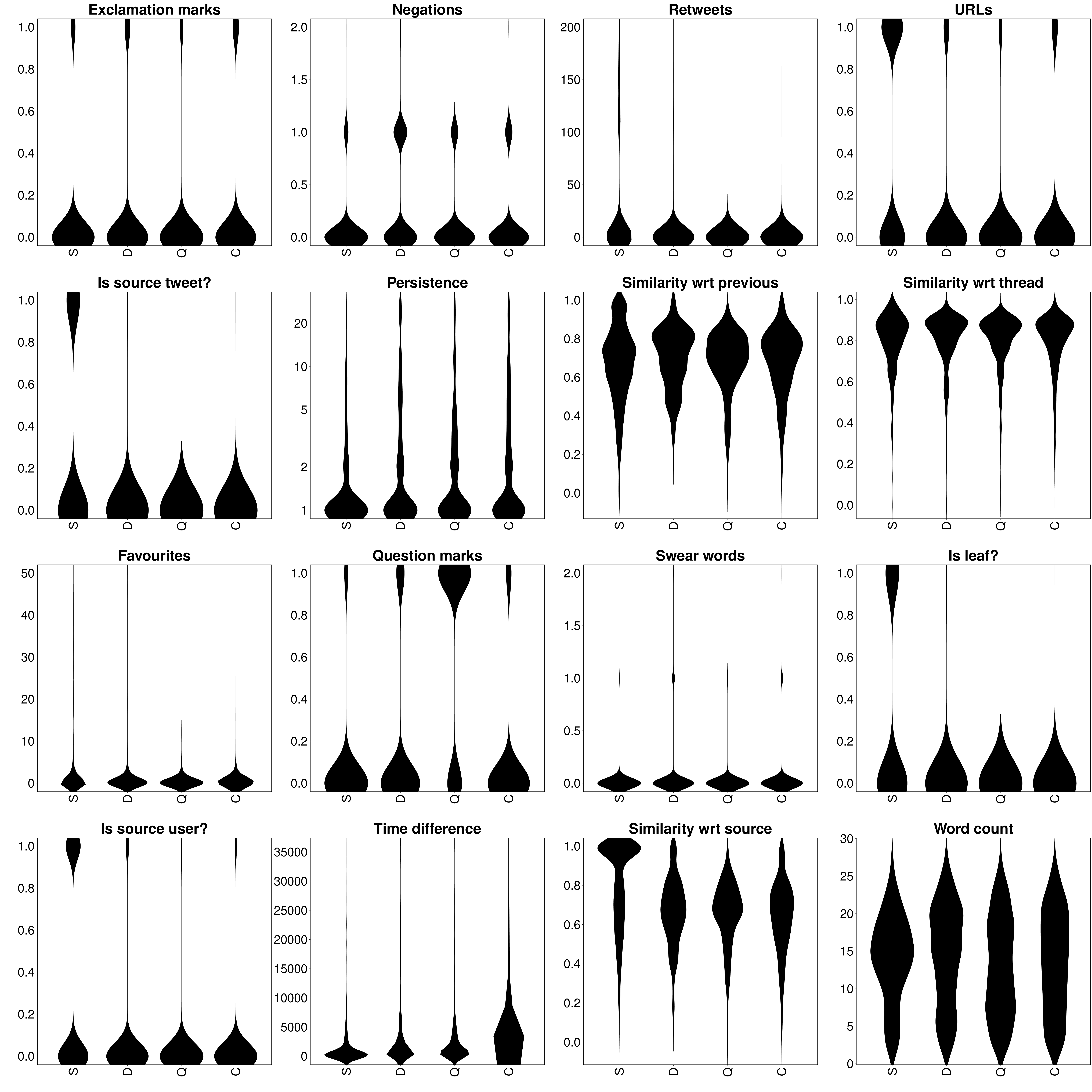}
 \caption{Distributions of feature values across the four categories: Support, Deny, Query and Comment.}
 \label{fig:features}
\end{figure}

To complete the analysis of our experiments, we now look at the different features we used in our study and perform an analysis to understand how distinctive the different features are for the four categories in the stance classification problem. We visualise the different distributions of features for the four categories in beanplots \cite{kampstra2008beanplot}. We show the visualisations pertaining to 16 of the features under study in Figure \ref{fig:features}. This analysis leads us to some interesting observations towards characterising the different types of stances:

\begin{itemize}
 \item As one might expect, \textit{querying tweets} are more likely to have question marks.
 \item Interestingly, \textit{supporting tweets} tend to have a higher similarity with respect to the source tweet, indicating that the similarity based on word embeddings can be a good feature to identify those tweets.
 \item \textit{Supporting tweets} are more likely to come from the user who posted the source tweet.
 \item \textit{Supporting tweets} are more likely to include links, which is likely indicative of tweets pointing to evidence that supports their position.
 \item Looking at the delay in time of different types of tweets (i.e., the \textit{time difference} feature), we see that \textit{supporting}, \textit{denying} and \textit{querying tweets} are more likely to be observed only in the early stages of a rumour, while later tweets tend to be mostly comments. In fact, these suggests that discussion around the veracity of a rumour occurs especially in the period just after it is posted, whereas the conversation then evolves towards comments that do not discuss the veracity of the rumour in question.
 \item \textit{Denying tweets} are more likely to use negating words. However, negations are also used in other kinds of tweets to a lesser extent, which also makes it more complicated for the classifiers to identify denying tweets. In addition to the low presence of denying tweets in the datasets, the use of negations also in other kinds of responses makes it more challenging to classify them. A way to overcome this may be to use more sophisticated approaches to identify negations that are rebutting the rumour initiated in the source tweet, while getting rid of the rest of the negations.
 \item When we look at the extent to which users persist in their participation in a conversational thread (i.e., the \textit{persistence} feature), we see that users tend to participate more when they are posting \textit{supporting tweets}, showing that users especially insistent when they support a rumour. However, we observe a difference that is not highly remarkable in this particular case.
\end{itemize}

The rest of the features do not show a clear tendency that helps visually distinguish characteristics of the different types of responses. While some features like swear words or exclamation marks may seem indicative of how they orient to somebody else's earlier post, there is no clear difference in reality in our datasets. The same is true for social features like retweets or favourites, where one may expect, for instance, that denying tweets may attract more retweets than comments, as people may want to let others know about rebuttals; the distributions of retweets and favourites are, however, very similar for the different categories.

One possible concern from this analysis is that there are very few features that characterise \textit{commenting tweets}. In fact, the only feature that we have identified as being clearly distinct for \textit{comments} is the \textit{time difference}, given that they are more likely to appear later in the conversations. This may well help classify those late \textit{comments}, however, early comments will be more difficult to be classified based on that feature. Finding additional features to distinguish \textit{comments} from the rest of the tweets may be of help for improving the overall classification.

\section{Conclusions and Future Work}

While discourse and sequential structure of social media conversations have been barely explored in previous work, our work has performed an analysis on the use of different sequential classifiers for the rumour stance classification task. Our work makes three core contributions to existing work on rumour stance classification: (1) we focus on the stance of tweets towards rumours that emerge while breaking news stories unfold; (2) we broaden the stance types considered in previous work to encompass all types of responses observed during breaking news, performing a 4-way classification task; and (3) instead of dealing with tweets as single units in isolation, we exploit the emergent structure of interactions between users on Twitter. In this task, a classifier has to determine if each tweet is supporting, denying, querying or commenting on a rumour's truth value. We mine the sequential structure of Twitter conversational threads in the form of users' replies to one another, extending existing approaches that treat each tweet as a separate unit. We have used four different sequential classifiers: (1) a Hawkes Process classifier that exploits temporal sequences, which showed state-of-the-art performance \cite{lukasik2016hawkes}; (2) a linear-chain CRF modelling tree-structured conversations broken down into branches; (3) a tree CRF modelling them as a graph that includes the whole tree; and (4) an LSTM classifier that also models the conversational threads as branches. These classifiers have been compared with a range of baseline classifiers, including the non-sequential equivalent Maximum Entropy classifier, on eight Twitter datasets associated with breaking news.

While previous stance detection work had mostly limited classifiers to looking at tweets as single units, we have shown that exploiting the discursive characteristics of interactions on Twitter, by considering probabilities of transitions within tree-structured conversational threads, can lead to substantial improvements. Among the sequential classifiers, our results show that the LSTM classifier using a more limited set of features performs the best, thanks to its ability to natively handle context, as well as only relying on branches instead of the whole tree, which reduces the amount of data and complexity that needs to be processed in each sequence. The LSTM has been shown to perform consistently well across datasets, as well as across different types of stances. Besides the comparison of classifiers, our analysis also looks at the distributions of the different features under study as well as how well they characterise the different types of stances. This enables us both to find out which features are the most useful, as well as to suggest improvements needed in future work for improving stance classifiers.

To the best of our knowledge, this is the first attempt at aggregating the conversational structure of Twitter threads to produce classifications at the tweet level. Besides the utility of mining sequences from conversational threads for stance classification, we believe that our results will, in turn, encourage the study of sequential classifiers applied to other natural language processing and data mining tasks where the output for each tweet can benefit from the structure of the entire conversation, e.g., sentiment analysis \cite{kouloumpis2011twitter,tsytsarau2012survey,saif2016contextual,liu2016identifying,vilares2017supervised,pandey2017twitter}, tweet geolocation \cite{han2014text,zubiaga2017towards}, language identification \cite{bergsma2012language,zubiaga2016tweetlid}, event detection \cite{srijith2017sub} and analysis of public perceptions on news \cite{reis2015breaking,an2011media} and other issues \cite{pak2010twitter,bian2016mining}.

Our plans for future work include further developing the set of features that characterise the most challenging and least-frequent stances, i.e., denying tweets and querying tweets. These need to be investigated as part of a more detailed and interdisciplinary, thematic analysis of threads \cite{tolmie2017microblog,housley2017digitizing,housley2017membership}. We also plan to develop an LSTM classifier that mines the entire conversation as a single tree. Our approach assumes that rumours have been already identified or input by a human, hence a final and ambitious aim for future work is the integration with our rumour detection system \cite{zubiaga2016learning}, whose output would be fed to the stance classification system. The output of our stance classification will also be integrated with a veracity classification system, where the aggregation of stances observed around a rumour will be exploited to determine the likely veracity of the rumour.

\section*{Acknowledgments}

This work has been supported by the PHEME FP7 project (grant No. 611233), the EPSRC Career Acceleration Fellowship EP/I004327/1, Elsevier through the UCL Big Data Institute, and The Alan Turing Institute under the EPSRC grant EP/N510129/1.

\section*{References}

\bibliographystyle{elsarticle-num}
\bibliography{sdqc}

\appendix{}

\section{Features}
\label{ap:features}

\subsection{Local Features}
\label{ssec:local-features}

Local features are extracted from each of the tweets in isolation, and therefore it is not necessary to look at other features in a thread to generate them. We use four types of features to represent the tweets locally.

\noindent \textbf{Local feature type \#1: Lexicon.}
\begin{itemize}
 \item \textit{Word Embeddings:} we use Word2Vec \cite{mikolov2013distributed} to represent the textual content of each tweet. First, we trained a separate Word2Vec model for each of the eight folds, each having the seven events in the training set as input data, so that the event (and the vocabulary) in the test set is unknown. We use large datasets associated with the seven events in the training set, including all the tweets we collected for those events. Finally, we represent each tweet as a vector with 300 dimensions averaging vector representations of the words in the tweet using Word2Vec.
 \item \textit{Part of speech (POS) tags:} we parse the tweets to extract the part-of-speech (POS) tags using Twitie \cite{bontcheva2013twitie}. Once the tweets are parsed, we represent each tweet with a vector that counts the number of occurrences of each type of POS tag. The final vector therefore has as many features as different types of POS tags we observe in the dataset.
 \item \textit{Use of negation:} this is a feature determining the number of negation words found in a tweet. The existence of negation words in a tweet is determined by looking at the presence of the following words: not, no, nobody, nothing, none, never, neither, nor, nowhere, hardly, scarcely, barely, don't, isn't, wasn't, shouldn't, wouldn't, couldn't, doesn't.
 \item \textit{Use of swear words:} this is a feature determining the number of `bad' words present in a tweet. We use a list of 458 bad words\footnote{\url{http://urbanoalvarez.es/blog/2008/04/04/bad-words-list/}}.
\end{itemize}

\noindent \textbf{Local feature type \#2: Content formatting.}
\begin{itemize}
 \item \textit{Tweet length:} the length of the tweet in number of characters.
 \item \textit{Word count:} the number of words in the tweet, counted as the number of space-separated tokens.
\end{itemize}

\noindent \textbf{Local feature type \#3: Punctuation.}
\begin{itemize}
 \item \textit{Use of question mark:} binary feature indicating the presence or not of at least one question mark in the tweet.
 \item \textit{Use of exclamation mark:} binary feature indicating the presence or not of at least one exclamation mark in the tweet.
\end{itemize}

\noindent \textbf{Local feature type \#4: Tweet formatting.}
\begin{itemize}
 \item \textit{Attachment of URL:} binary feature, capturing the presence or not of at least one URL in the tweet.
\end{itemize}

\subsection{Contextual Features}
\label{ssec:contextual-features}

\noindent \textbf{Contextual feature type \#1: Relational features.}

\begin{itemize}
 \item \textit{Word2Vec similarity wrt source tweet:} we compute the cosine similarity between the word vector representation of the current tweet and the word vector representation of the source tweet. This feature intends to capture the semantic relationship between the current tweet and the source tweet and therefore help inferring the type of response.
 \item \textit{Word2Vec similarity wrt preceding tweet:} likewise, we compute the similarity between the current tweet and the preceding tweet, the one that is directly responding to.
 \item \textit{Word2Vec similarity wrt thread:} we compute another similarity score between the current tweet and the rest of the tweets in the thread excluding the tweets from the same author as that in the current tweet.
\end{itemize}

\noindent \textbf{Contextual feature type \#2: Structural features.}

\begin{itemize}
 \item \textit{Is leaf:} binary feature indicating if the current tweet is a leaf, i.e. the last tweet in a branch of the tree, with no more replies following.
 \item \textit{Is source tweet:} binary feature determining if the tweet is a source tweet or is instead replying to someone else. Note that this feature can also be extracted from the tweet itself, checking if the tweet content begins with a Twitter user handle or not.
 \item \textit{Is source user:} binary feature indicating if the current tweet is posted by the same author as that in the source tweet.
\end{itemize}

\noindent \textbf{Contextual feature type \#3: Social features.}

\begin{itemize}
 \item \textit{Has favourites:} feature indicating the number of times a tweet has been favourited.
 \item \textit{Has retweets:} feature indicating the number of times a tweet has been retweeted.
 \item \textit{Persistence:} this feature is the count of the total number of tweets posted in the thread by the author in the current tweet. High numbers of tweets in a thread indicate that the author participates more.
 \item \textit{Time difference:} this is the time elapsed, in seconds, from when the source tweet was posted to the time the current tweet was posted.
\end{itemize}

\subsection{Hawkes Features}
\label{ssec:hawkes-features}

\begin{itemize}
  \item \textit{Bag of words:} a vector where each token in the dataset represents a feature, where each feature is assigned a number pertaining its count of occurrences in the tweet.
  \item \textit{Timestamp:} The UNIX time in which the tweet was posted.
\end{itemize}

\end{document}